\documentclass[runningheads]{llncs}
\usepackage{graphicx}
\usepackage[colorlinks=true, citecolor=black, linkcolor=black]{hyperref}

\usepackage{listings}
\usepackage[caption=false]{subfig}

\newcommand{\reffigure}[1]{\figurename~\ref{#1}}

\usepackage{algorithm}
\usepackage{algpseudocode}
\usepackage{booktabs}



\begin{document}
\title{Automatic Generation of Board Game Manuals}
%
%

\author{Matthew Stephenson \and {\'E}ric Piette \and Dennis J. N. J. Soemers \and Cameron Browne}
\authorrunning{M. Stephenson et al.}
%
\institute{Department of Data Science and Knowledge Engineering, Maastricht University, Paul-Henri Spaaklaan 1, 6229 EN, Maastricht, the Netherlands
\email{\{matthew.stephenson,eric.piette,dennis.soemers,cameron.browne\}\\@maastrichtuniversity.nl}}
\maketitle              
\begin{abstract}
In this paper we present a process for automatically generating manuals for board games within the Ludii general game system. This process requires many different sub-tasks to be addressed, such as English translation of Ludii game descriptions, move visualisation, highlighting winning moves, strategy explanation, among others. These aspects are then combined to create a full manual for any given game. This manual is intended to provide a more intuitive explanation of a game's rules and mechanics, particularly for players who are less familiar with the Ludii game description language and grammar.

\end{abstract}
%
%
%


\section{Introduction}
Board games are one of the most popular pastimes for millions of people, and have been played for over 5000 years \cite{Browne_2018_Modern}. Board Game Geek, one of the most popular repositories of board games, currently includes over 130,000 different games.\footnote{\url{https://boardgamegeek.com}}
As such, manually describing the rules for all these games in a clear and consistent manner is a near impossible task. In this paper we present a game manual generation framework, which automatically creates explanations for how to play any given board game. This framework operates on the Ludii general game system, which supports the majority of non-dexterity board games. 

The rest of this paper is organised as follows. Section 2 provides the necessary background information about Ludii and other related work. Section 3 describes the Ludii game description language and our process for translating it into English. Section 4 describes how different moves for a given game are detected, classified and visualised. Section 5 describes several additional features that can provide supplementary information. Section 6 describes how these previous aspects are combined into a complete game manual. Section 7 concludes this work, discusses several limitations, and suggests ideas for future research.


\section{Background}

\subsection{Ludii}
Ludii is a general game system \cite{Piette_2020_Ludii} that is being developed as part of the Digital Ludeme Project \cite{Browne_2018_Modern}. The majority of games within Ludii are traditional board games, but other types of games such as puzzles, finger games, dice games, etc., are also present. Ludii is also able to support games which are stochastic or contain hidden information. Ludii currently includes over 950 playable games,\footnote{\url{https://ludii.games/download}, version 1.2.9} covering a wide-range of different categories and mechanics.

\subsection{Related Work}

\subsubsection{Video Games}
The work that most closely resembles our desired outcome would probably be the AtDelfi system \cite{Green_2018_AtDELFI} for the General Video Game AI (GVGAI) framework \cite{Perez_2016_GVGAI,Perez_2019_GVGAI}. This system generates instruction cards for simple arcade-style video games written in the video game description language (VGDL), providing information about the game's controls, how to gain points, and how the game is won or lost. This information is displayed using a combination of text, images, and GIF animations. Although this approach focuses on video games rather than board games, the generated instruction cards are similar in design and purpose to the manuals we would like to produce.

However, there are several additional considerations when creating manuals for board games. Board games often have very different control schemes to those of video games, where moves are defined less by the exact buttons the player can press and more by what pieces can be placed or moved in accordance with the game's rules. Video games also typically have specific considerations which do not apply to board games, such as non-player characters (NPCs), projectiles, timed events, a larger state space, etc. Likewise, board games often contain aspects such as multiple players, distinct game-play phases, a greater reliance on strategy rather than dexterity, a larger action space, etc. Because of this, our generated board game manuals will likely contain very different information from AtDelfi, with a much greater emphasis on piece movement and the specific rules of play.

\subsubsection{Learning From Observation}
This work involves learning the rules of board games from observed play \cite{Bjornsson_2012_Learning,Gregory_2016_GRLSystem,Cropper_2020_Inductive,Kowalski_2018_Regular}, where a learner agent is given a collection of play traces and tasked with learning the rules of the game which produced them.
Prior work in this area has predominately focused on learning from games written in the Stanford Game Description Language (GDL), which is a much more verbose and low-level language compared to that of Ludii. This makes direct translation from the game description very difficult, hence the reliance on learning from observed moves. 
Research on this task is also somewhat limited, with presented approaches so far only being able to translate a limited subset of GDL game descriptions.

Our presented approach instead relies more heavily on the higher level language provided by Ludii, utilising the ability to specify English translations for specific sub-sections of a games's description based on its wider context. This arguably makes our approach less general than learning from observation, as it is currently unable to operate on games outside of Ludii. However, our approach works effectively on the majority of board games within Ludii, and the presented ideas are likely to be applicable to other high-level game description languages with only minor modifications.


\section{Ludii Game Description Language}
Every game within Ludii is described as a single symbolic expression, which contains structured sets of pre-defined keywords and values. These keywords are called ludemes, and are intended to represent some fundamental aspect of a game (e.g. \texttt{Line}, \texttt{Mover}, \texttt{Win}). Some ludemes are atomic and require no additional parameters, such as the examples given above, while others such as \texttt{if} require additional arguments to be provided, in this case a condition and statement.
By combining several ludemes and values together into compound expressions, we can create larger ludemeplexes that describe more complex ideas. For example, \texttt{(if (is Line 3) (result Mover Win))} describes that forming a line of 3 pieces results in a win. This same idea is used to describe all of the games in Ludii, repeatedly creating and combining increasingly complex ludemeplexes until the final game description is obtained. As an example, the complete Ludii game description for Tic-Tac-Toe is shown in \reffigure{Fig:TicTacToeDescription}.

\begin{figure}[!t]
\centering
{
\begin{lstlisting}
    (game "Tic-Tac-Toe"  
        (players 2)  
        (equipment { 
            (board (square 3)) 
            (piece "Disc" P1) 
            (piece "Cross" P2) 
        })  
        (rules 
            (play (move Add (to (sites Empty))))
            (end (if (is Line 3) (result Mover Win)))
        )
    )
\end{lstlisting}
}
\caption{Game description for Tic-Tac-Toe, written in Ludii's game description language.}
\label{Fig:TicTacToeDescription}
\end{figure}

The number of ludemes needed to describe a game varies based on its complexity, typically ranging from only a few dozen for simple examples like Tic-Tac-Toe and Hex, to several hundred for more complex cases like Chess and Backgammon. The complete set of ludemes within Ludii is referred to as the Ludii Game Description Language (L-GDL), and is automatically inferred from the Ludii codebase using a class grammar approach \cite{Browne_2016_Class}. Further details on L-GDL are provided in the Ludii game logic guide \cite{Piette_2021_LudiiGameLogicGuide}.

\subsection{Translation to English}
While Ludii game descriptions are generally clear and understandable for simple cases like Tic-Tac-Toe, deciphering more complex game descriptions often requires expert knowledge about L-GDL. As a result, Ludii relies on hand-written descriptions to accompany each game. These are often obtained from Board Game Geek, Wikipedia, or other sources with inconsistent levels of detail and structure. Even worse, some of these descriptions might make allusions to other games which players may be unfamiliar with, e.g. ``Knights move the same as in Chess''. All these issues motivate the creation of an automated game-translation process, which is able to convert any Ludii game description into a pseudo-English equivalent.

Due to the class grammar approach used to generate L-GDL, each Ludeme has a corresponding class within the Ludii codebase. Within each of these classes we define a toEnglish() function, which returns an English language description of how the ludeme operates within the game. These descriptions often need to consider the ludeme's arguments, for example \texttt{(is Line 3)} might translate to ``3 pieces in a line''. As such, the toEnglish() function for each ludeme will often need to call the toEnglish() functions of its arguments. For example, the toEnglish() function of the \texttt{if} ludeme also calls the toEnglish() of its conditional and statement arguments. This structure means that calling the toEnglish() function on the highest level \texttt{(game ...)} ludeme of any given game description, will recursively call and combine the results of all its sub-ludemes to create a full English language translation of the game. As an example, applying this process on the description in \reffigure{Fig:TicTacToeDescription}, outputs the text shown in \reffigure{Fig:TicTacToeDescriptionEnglish}. Additional examples of English translations of Ludii game descriptions are shown in Appendix \ref{appendixA}.

\begin{figure}[!t]
\centering
{
\begin{lstlisting}
The game "Tic-Tac-Toe" is played by two players on a 3x3 rectangle board with square tiling. 
Player one plays with Discs. Player two plays with Crosses.
Players take turns moving.
Rules: 
     Add one of your pieces to the set of empty cells.
Aim: 
     If a player places 3 of their pieces in an adjacent direction line, the moving player wins.
\end{lstlisting}
}
\caption{Automatic English language translation of Ludii's Tic-Tac-Toe game description.}
\label{Fig:TicTacToeDescriptionEnglish}
\end{figure}

However, due to the complexities and inconsistencies of the English language, this translation process is unlikely to work perfectly for every case. This includes mostly harmless grammar issues like displaying the correct pluralisation of different words, to more complex challenges like converting a sequence of nested statements such as \texttt{(or A (or (or B C) D) (or E F))} into an unambiguous sentence. 

Ludii also contains certain implementation-specific ludemes such as \textit{SetPending} or \textit{SetState}, which are often used for different purposes across multiple games. For example, in the game Chess the state of a piece indicates whether it has previously been moved or not, and is used to determine if castling is possible. Alternatively, in the game Jungle the state of a piece indicates its combat strength, and is used to determine if certain pieces can capture others. As of the time of writing, our translation approach is currently unable to directly link the action of setting a piece's state with the possible future consequences of this.


\section {Move Visualisations}
While the English translation of a game's description provides an explanation of the possible moves a player can make, there are some moves which are much easier to understand visually. Chess provides a nice example of this, where showing the movements of the different pieces such as Knight, Queen, Rook, and Bishop using diagrams or animations may be much more intuitive than with text alone. Therefore, we also developed a move visualisation process which attempts to identify all of the different types of moves that each player/piece can perform, and creates suitable images to demonstrate them.

\subsection{Move Properties}
\label{move_properties}
For the purposes of move visualisation, there are four move properties that need to be considered: 
\begin{enumerate}
    \item \textbf{Mover:} Each move contains a \textit{Mover} parameter, which indicates the player who is making the move. This property is only considered for games where the players have different piece rules.
    \item \textbf{Piece:} The majority of moves contain an associated \textit{Piece} parameter, which indicates the main piece that is moved, added, or removed. Even though a move may affect multiple pieces, a single piece is always designated as the main one. For example, when capturing a piece in chess there are technically two pieces that are being affected (the capturing piece and the captured piece), but the capturing piece is considered the main piece. Certain moves such as \textit{Pass} or \textit{Swap} do not have an associated piece.
    \item \textbf{Origin Rules:} Each move originates from an associated \texttt{(move ...)} ludeme within the current game's description. For example, when playing the game description given in \reffigure{Fig:TicTacToeDescription}, all moves originate from \texttt{(move Add (to (sites Empty)))}. By calling our previously described toEnglish() function on any move's associated ludeme, we can obtain an English translation of the rules from which this move originated.
    \item \textbf{Action Types:} Each move contains a sequence of actions that are applied when the move is selected. For example, the move \texttt{[(Remove E6), (Move F5-E6), (Score P1=4)]} contains three actions which removes the piece at E6, moves the piece at F5 to E6, and sets the score of Player 1 to 4, respectively. The action types of a move are this same set of actions but where only the name/type of action is retained; for example the action types of the same example move given above would be \texttt{[Remove, Move, Score]}.
\end{enumerate}

\subsection{Identifying Distinct Moves}
When performing move visualisation for a given game, we first run a number of random playouts. Once completed, we then combine all of the moves that were selected during these playouts into a single list of moves. We then remove all duplicate moves from this list, using only the four move properties listed in Section \ref{move_properties} to determine uniqueness. This provides us with a set of distinct moves, each of which has a unique combination of properties.

\subsection{Visual Representations}
\begin{figure}[!t]
     \centering
     \subfloat[Before Move]{%
        \includegraphics[width=.4\textwidth]{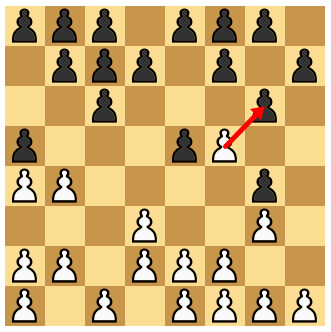}
    }
     ~~~~~~~~
     \subfloat[After Move]{%
        \includegraphics[width=.405\textwidth]{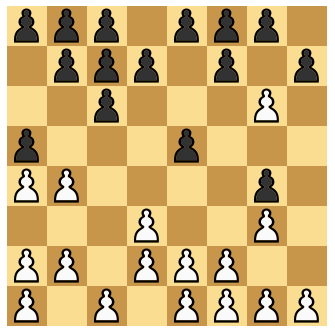}
    }
        \caption{Visual representation of a move for Breakthrough, showing the board state before (a) and after (b) the move is selected. The move is highlighted using a red arrow.}
        \label{Fig:example_move_images}
\end{figure}

For each distinct move, we create two images showing the state of the game before and after the move is selected. The move's piece is also highlighted using either a red arrow or dot, depending on whether the piece's location changes or not. An example of these two images for a given move is shown in \reffigure{Fig:example_move_images}. 
Additionally, certain games within Ludii support move animations which can depict a moving piece in a more visually pleasing manner. In games where this animation is supported, we also provide a short GIF animation of the move.

\section {Additional Features}

As well as the English translation and move visualisation processes, there are several auxiliary features which can provide additional information.

\subsection{Initial Setup}
An image of the game state before any moves are made is also included. This is not very helpful for games where the board is empty, but can be beneficial for games such as Chess where the initial arrangement of the pieces is important.

\subsection{Winning / Losing Moves}
When running random playouts for move visualisation, we can also record the result of each playout as well as the final move that was made.
We can then visualise this move for each unique result, showing the different possible results that the game can have as well as the last move that led to this outcome.

\begin{figure}[!t]
     \centering
     \subfloat[Before Move]{%
        \includegraphics[width=.4\textwidth]{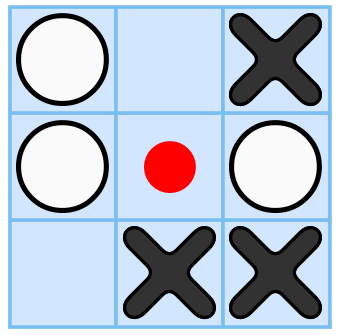}
    }
      ~~~~~~~~
     \subfloat[After Move]{%
        \includegraphics[width=.4\textwidth]{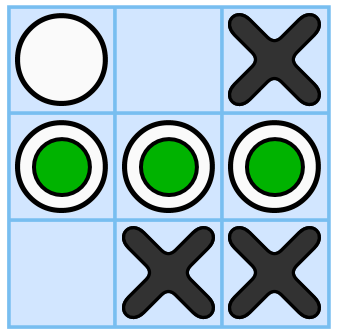}
    }
        \caption{Visual representation of an ending move for Tic-Tac-Toe, showing the board state before (a) and after (b) the move is selected. The move is highlighted using a red dot, and the winning line is highlighted using green dots.}
        \label{Fig:example_end_images}
\end{figure}

In addition to the regular move visualisation, we can also detect the specific \texttt{(end ...)} ludeme that caused the game to end. By calling our previously described toEnglish() function on this ludeme, we can obtain an English translation of the rules that ended the game. Certain \texttt{(end ...)} ludemes also support additional visuals, such as allowing us to identify the specific winning line or pattern. An example of these two ending move images for a result in Tic-Tac-Toe is shown in \reffigure{Fig:example_end_images}. 

\subsection{Similar Legal Moves}
\begin{figure}[!t]
     \centering
     \subfloat[Only Selected Move]{%
        \includegraphics[width=.4\textwidth]{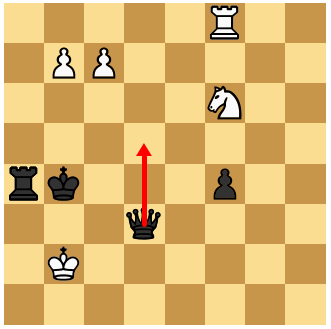}
    }
      ~~~~~~~~
     \subfloat[All Similar Moves]{%
        \includegraphics[width=.4\textwidth]{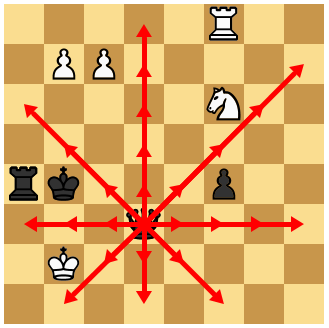}
    }
        \caption{Visual representation of the board state for Chess before a move is made, with only the selected move highlighted (a) and all legal moves with the same properties highlighted (b).}
        \label{Fig:additional_move_images}
\end{figure}

When highlighting moves using red arrows or dots, we can also highlight any other legal moves that have the same properties. Whereas previously only the selected move was shown, this approach displays more example moves within a single picture. Contrasting images showing with and without this addition are shown in \reffigure{Fig:additional_move_images}. Both approaches have their benefits, however we feel that showing all similar legal moves more closely aligns with how piece rules are typically described in most Ludii game descriptions. 

\subsection{Strategy Explanation}
Many of the games within Ludii also contain associated heuristics for assisting AI agents. The heuristics include aspects such as Material (number of pieces), Mobility (number of moves), LineCompletion (potential ability to complete lines of pieces), among many others. Each heuristic also has an associated weight which indicates its relative level of importance. These heuristics can be specified manually by the game's designer, or learned automatically through a heuristic tuning process. More details on the heuristics which are available within Ludii can be found in \cite{Stephenson_2021_General}. While predominately designed for AI agents, these heuristics can also provide assistance to novice players about what moves to make. By applying some simple formatting, these heuristics can be converted into basic strategy explanations. An example of this strategy explanation for the Chess heuristics in Ludii is shown in \reffigure{Fig:strategyExplanation}.

\begin{figure}[!t]
\centering
{
\begin{lstlisting}
Try to maximise the number of Pawn(s) you control (very low importance)
Try to maximise the number of Rook(s) you control moderate importance)
Try to maximise the number of Bishop(s) you control (low importance)
Try to maximise the number of Knight(s) you control (low importance)
Try to maximise the number of Queen(s) you control (very high importance)
\end{lstlisting}
}
\caption{Strategy explanation derived from Ludii's AI heuristics for Chess.}
\label{Fig:strategyExplanation}
\end{figure}


\section {Complete game Manuals}
All of the aspects described in Sections 3, 4, and 5 can be combined into a single webpage document which provides complete manuals for a specific game. Examples of generated manuals for a wide range of Ludii games are available online.\footnote{\url{https://ludii.games/manuals/menu.html}}
The layout of each manual is as follows:
\begin{enumerate}
    \item \textbf{Rules:} The rules of the game, based on the English translation of its Ludii game description.
    \item \textbf{Heuristics:} Recommended strategies for playing the game, derived from the AI heuristics.
    \item \textbf{Setup:} The initial state of the game before any moves are made.
    \item \textbf{Endings:} The different endings for the game, with accompanying English descriptions and visualisations.
    \item \textbf{Moves:} The different moves for the game, with accompanying English descriptions and visualisations. These moves are hierarchically organised based on their properties (mover, piece, origin rules, and action types).
\end{enumerate}


\section{Conclusion}
In this paper we have presented a process for automatically generating manuals of Ludii games. This process combines solutions for multiple sub-tasks to create a complete manual, detailing the rules, moves, endings and strategies for any given game. This manual allows players who are unfamiliar with L-GDL to still play complex games in Ludii, even if the hand-written rules are unavailable or incomplete. Procedurally generated games would be an ideal application of this work, as these games do not provide any instructions or details beyond their Ludii game description. These manuals may also be able to assist game designers, by providing a starting point for describing the game's rules or highlighting outlier moves that the designer may not have intended.

\subsection{Limitations}
While the manual generation process described in this paper works on the vast majority of games in Ludii, there are still some minor cases where specific parts are incompatible. Move visualisation is currently not possible for games with simultaneous moves, or games which reference smaller sub-games (aka. matches). The English translation code is also still in-progress, and will need to be updated and improved over time. Other minor aspects such as move animations and winning move visuals are also not yet implemented for all cases.

\subsection{Future Work}
One improvement on this work could be the creation of tutorial scenarios where a specific mechanic or rule needs to be understood in order to win, thus providing a more interactive learning experience \cite{Green_2018_Generating}. Interactive tutorials have been shown to increase player engagement and ability, particularly for complex games \cite{Andersen_2012_Impact,Kelleher_2005_Stencils}. These scenarios could be created by identifying game states where a certain move needs to be selected in order to win, or avoid loss.


Another improvement could be the addition of a coach/tutor AI that provides tips to the player throughout the game. This could include advice about what moves to make, or feedback on why certain moves were good/bad \cite{Ikeda_2016_Detection}. This could be achieved using our learned heuristics, by detecting if a different move would have given a higher state evaluation. For example, ``Your last move resulted in a lower material benefit (+2) than an alternative move (+5)''.
Furthermore, features of state-action pairs \cite{Browne_2019_Strategic} may be used to provide advice on a tactical level, as opposed to the strategic level that heuristic state functions operate on.

Our final improvement suggestion, is the creation of an adaptive AI opponent which modifies its strength based on the player's abilities \cite{Iida_1995_Tutoring}. This agent would aim to provide a reasonable level of challenge for the player, hopefully providing a more engaging and constructive gameplay experience. This agent could be extended further to favour states which promote certain strategies, providing an online version of the tutorial scenarios mentioned previously.


\section*{Acknowledgements}

This research is funded by the European Research Council as part of the Digital Ludeme Project (ERC Consolidator Grant \#771292) led by Cameron Browne at Maastricht University's Department of Data Science and Knowledge Engineering.

\bibliographystyle{splncs04}
\bibliography{dlp-biblio-1}

\begin{thebibliography}{10}
\providecommand{\url}[1]{\texttt{#1}}
\providecommand{\urlprefix}{URL }
\providecommand{\doi}[1]{https://doi.org/#1}

\bibitem{Andersen_2012_Impact}
Andersen, E., O'Rourke, E., Liu, Y.E., Snider, R., Lowdermilk, J., Truong, D.,
  Cooper, S., Popovic, Z.: The Impact of Tutorials on Games of Varying
  Complexity, p. 59–68. Association for Computing Machinery, New York, NY,
  USA (2012)

\bibitem{Bjornsson_2012_Learning}
Bj\"{o}rnsson, Y.: Learning rules of simplified boardgames by observing. In:
  Proceedings of the Twentieth European Conference on Artificial Intelligence.
  pp. 175--180 (2012)

\bibitem{Browne_2018_Modern}
Browne, C.: Modern techniques for ancient games. In: IEEE Conference on
  Computational Intelligence and Games. pp. 490--497. IEEE Press, Maastricht
  (2018)

\bibitem{Browne_2019_Strategic}
Browne, C., Soemers, D.J.N.J., Piette, E.: Strategic features for general
  games. In: Proceedings of the 2nd Workshop on Knowledge Extraction from Games
  (KEG). pp. 70--75 (2019)

\bibitem{Browne_2016_Class}
Browne, C.B.: A class grammar for general games. In: Advances in Computer
  Games. Lecture Notes in Computer Science, vol. 10068, pp. 167--182. Leiden
  (2016)

\bibitem{Cropper_2020_Inductive}
Cropper, A., Evans, R., Law, M.: Inductive general game playing. Machine
  Learning  \textbf{109}(7),  1393--1434 (2020)

\bibitem{Green_2018_AtDELFI}
Green, M.C., Khalifa, A., Barros, G.A.B., Machado, T., Nealen, A., Togelius,
  J.: Atdelfi: Automatically designing legible, full instructions for games.
  In: Proceedings of the 13th International Conference on the Foundations of
  Digital Games. FDG '18, Association for Computing Machinery, New York, NY,
  USA (2018)

\bibitem{Green_2018_Generating}
Green, M.C., Khalifa, A., Barros, G.A.B., Nealen, A., Togelius, J.: Generating
  levels that teach mechanics. In: Proceedings of the 13th International
  Conference on the Foundations of Digital Games. FDG '18, Association for
  Computing Machinery, New York, NY, USA (2018)

\bibitem{Gregory_2016_GRLSystem}
{Gregory}, P., {Schumann}, H., {Bj\"{o}rnsson}, Y., {Schiffel}, S.: The {GRL}
  system: Learning board game rules with piece-move interactions. In: Computer
  Games. pp. 130--148. Springer International Publishing (2016)

\bibitem{Iida_1995_Tutoring}
{Iida}, H., {Matsubara}, H., {Uiterwijk}, J.: A search strategy for tutoring in
  game playing. IJCAI-95 Workshop Proc., Entertainment and AI/Alife pp. 14--18
  (1995)

\bibitem{Ikeda_2016_Detection}
Ikeda, K., Viennot, S., Sato, N.: Detection and labeling of bad moves for
  coaching {G}o. In: 2016 IEEE Conference on Computational Intelligence and
  Games (CIG). pp. 395--401 (2016)

\bibitem{Kelleher_2005_Stencils}
Kelleher, C., Pausch, R.: Stencils-Based Tutorials: Design and Evaluation, p.
  541–550. Association for Computing Machinery, New York, NY, USA (2005)

\bibitem{Kowalski_2018_Regular}
Kowalski, J., Kisielewicz, A.: Regular language inference for learning rules of
  simplified boardgames. In: 2018 IEEE Conference on Computational Intelligence
  and Games (CIG). pp. 78--85 (2018)

\bibitem{Perez_2019_GVGAI}
{Perez-Liebana}, D., {Liu}, J., {Khalifa}, A., {Gaina}, R.D., {Togelius}, J.,
  {Lucas}, S.M.: General video game ai: A multitrack framework for evaluating
  agents, games, and content generation algorithms. IEEE Transactions on Games
  \textbf{11}(3),  195--214 (2019)

\bibitem{Perez_2016_GVGAI}
Perez-Liebana, D., Samothrakis, S., Togelius, J., Lucas, S.M., Schaul, T.:
  General video game ai: Competition, challenges, and opportunities. In:
  Proceedings of the 30th AAAI Conference on Artificial Intelligence. pp.
  4335--4337 (2016)

\bibitem{Piette_2021_LudiiGameLogicGuide}
Piette, {\'E}., Browne, C., Soemers, D.J.N.J.: Ludii game logic guide.
  \url{https://arxiv.org/abs/2101.02120} (2021)

\bibitem{Piette_2020_Ludii}
Piette, {\'E}., Soemers, D.J.N.J., Stephenson, M., Sironi, C.F., Winands,
  M.H.M., Browne, C.: Ludii -- the ludemic general game system. In: Giacomo,
  G.D., Catala, A., Dilkina, B., Milano, M., Barro, S., Bugarín, A., Lang, J.
  (eds.) Proceedings of the 24th European Conference on Artificial Intelligence
  (ECAI 2020). Frontiers in Artificial Intelligence and Applications, vol.~325,
  pp. 411--418. IOS Press (2020)

\bibitem{Stephenson_2021_General}
Stephenson, M., Soemers, D.J.N.J., Piette, E., Browne, C.: General game
  heuristic prediction based on ludeme descriptions.
  \url{https://arxiv.org/abs/2105.12846} (2021)

\end{thebibliography}

\newpage

\section*{Appendix} 
\appendix

\section{English translations of Ludii game descriptions}
\label{appendixA}


\subsection{Hex}
\scriptsize
\begin{lstlisting}
(game "Hex"
    (players 2)
    (equipment{
            (board (hex Diamond 11))
            (piece "Marker" Each)
            (regions P1 { (sites Side NE) (sites Side SW) })
            (regions P2 { (sites Side NW) (sites Side SE) })})
    (rules
        (meta (swap))
        (play (move Add (to (sites Empty))))
        (end (if (is Connected Mover) (result Mover Win))))
)


The game "Hex" is played by two players on a 11x11 diamond board with hexagonal tiling.
Regions:
    RegionP1: the NE side for P1 and RegionP1: the SW side for P1
    RegionP2: the NW side for P2 and RegionP2: the SE side for P2 
All players play with Markers.
Players take turns moving.
Rules: 
     Add one of your pieces to the set of empty cells.
Aim: 
     If the region(s) of the moving player are connected, the moving player wins.
\end{lstlisting}

\subsection{Amazons}
\scriptsize
\begin{lstlisting}
(game "Amazons"
    (players 2)
    (equipment{
            (board (square 10))
            (piece "Queen" Each (move Slide (then (moveAgain))))
            (piece "Dot" Neutral)})
    (rules
        (start{
                (place "Queen1" {"A4" "D1" "G1" "J4"})
                (place "Queen2" {"A7" "D10" "G10" "J7"})})
        (play
            (if (is Even (count Moves))
                (forEach Piece)
                (move Shoot (piece "Dot0"))))
        (end (if (no Moves Next) (result Mover Win))))
)


The game "Amazons" is played by two players on a 10x10 rectangle board with square tiling. 
All players play with Queens. The following pieces are neutral: Dots.
Rules for Pieces:
     Queens slide from the location of the piece in the adjacent direction through the set of empty cells then move again.
Players take turns moving.
Setup:
     Place a Queen for player one on sites: A4, D1, G1 and J4.
     Place a Queen for player two on sites: A7, D10, G10 and J7.
Rules: 
     If the number of moves is even, move one of your pieces, else shoot the piece Dot0.
Aim: 
     If the next player cannot move, the moving player wins.
\end{lstlisting}

\end{document}